\def\year{2022}\relax
\documentclass[letterpaper]{article} % DO NOT CHANGE THIS
\usepackage{aaai22}  % DO NOT CHANGE THIS
\usepackage{times}  % DO NOT CHANGE THIS
\usepackage{helvet}  % DO NOT CHANGE THIS
\usepackage{courier}  % DO NOT CHANGE THIS
\usepackage[hyphens]{url}  % DO NOT CHANGE THIS
\usepackage{graphicx} % DO NOT CHANGE THIS
\usepackage{natbib}  % DO NOT CHANGE THIS AND DO NOT ADD ANY OPTIONS TO IT
\usepackage{caption} % DO NOT CHANGE THIS AND DO NOT ADD ANY OPTIONS TO IT
\DeclareCaptionStyle{ruled}{labelfont=normalfont,labelsep=colon,strut=off} % DO NOT CHANGE THIS
\usepackage{algorithm}
\usepackage{algorithmic}

\usepackage{newfloat}
\usepackage{listings}
\floatstyle{ruled}
\newfloat{listing}{tb}{lst}{}
\floatname{listing}{Listing}
\usepackage{amsmath,amsthm,mathtools}
\usepackage{diagbox}
\usepackage{subfigure}
\usepackage{multirow}
\usepackage{url}

\title{Structured Semantic Transfer for Multi-Label Recognition with Partial Labels}
\author{
    Tianshui Chen \textsuperscript{\rm 1},
    Tao Pu \textsuperscript{\rm 2}, 
    Hefeng Wu  \textsuperscript{\rm 2},
    Yuan Xie \textsuperscript{\rm 2}, 
    Liang Lin  \textsuperscript{\rm 2} \thanks{Tianshui Chen and Tao Pu contribute equally to this work and share first authorship. Corresponding author is Liang Lin.}
}
\affiliations{
    \textsuperscript{\rm 1} Guangdong University of Technology, 
    \textsuperscript{\rm 2} Sun Yat-Sen University\\
    tianshuichen@gmail.com, putao3@mail2.sysu.edu.cn, wuhefeng@gmail.com, phoenixsysu@gmail.com, linliang@ieee.org
}

\usepackage{bibentry}
\begin{document}

\maketitle

\begin{abstract}
Multi-label image recognition is a fundamental yet practical task because real-world images inherently possess multiple semantic labels. However, it is difficult to collect large-scale multi-label annotations due to the complexity of both the input images and output label spaces. To reduce the annotation cost, we propose a structured semantic transfer (SST) framework that enables training multi-label recognition models with partial labels, i.e., merely some labels are known while other labels are missing (also called unknown labels) per image. The framework consists of two complementary transfer modules that explore within-image and cross-image semantic correlations to transfer knowledge of known labels to generate pseudo labels for unknown labels. Specifically, an intra-image semantic transfer module learns image-specific label co-occurrence matrix and maps the known labels to complement unknown labels based on this matrix. Meanwhile, a cross-image transfer module learns category-specific feature similarities and helps complement unknown labels with high similarities. Finally, both known and generated labels are used to train the multi-label recognition models. Extensive experiments on the Microsoft COCO, Visual Genome and Pascal VOC datasets show that the proposed SST framework obtains superior performance over current state-of-the-art algorithms. Codes are available at \url{https://github.com/HCPLab-SYSU/HCP-MLR-PL}.
\end{abstract}

\section{Introduction}
Recently, lots of efforts \cite{chen2019multi,chen2019learning,chen2020knowledge} are dedicated to the task of multi-label image recognition as it benefits various applications ranging from content-based image retrieval and recommendation systems to surveillance systems and assistive robots. Despite achieving impressive progress, current leading algorithms \cite{chen2019multi,chen2019learning,chen2020knowledge} introduce data-hungry deep convolutional networks \cite{he2016deep,simonyan2015very} to learn discriminative features, and thus they depend on collecting large-scale clean and complete multi-label datasets. However, it is very time-consuming to collect a consistent and exhaustive list of labels for every image, making collecting clean and complete multi-label annotations more difficult and less scalable. In contrast, it is easy and scalable to annotate partial labels for each image, which can be regarded as an alternative way to address the above problem. In this work, we aim to address the task of learning multi-label recognition models with partial labels (MLR-PL).

\begin{figure}[!t]
   \centering
   \includegraphics[width=0.85\linewidth]{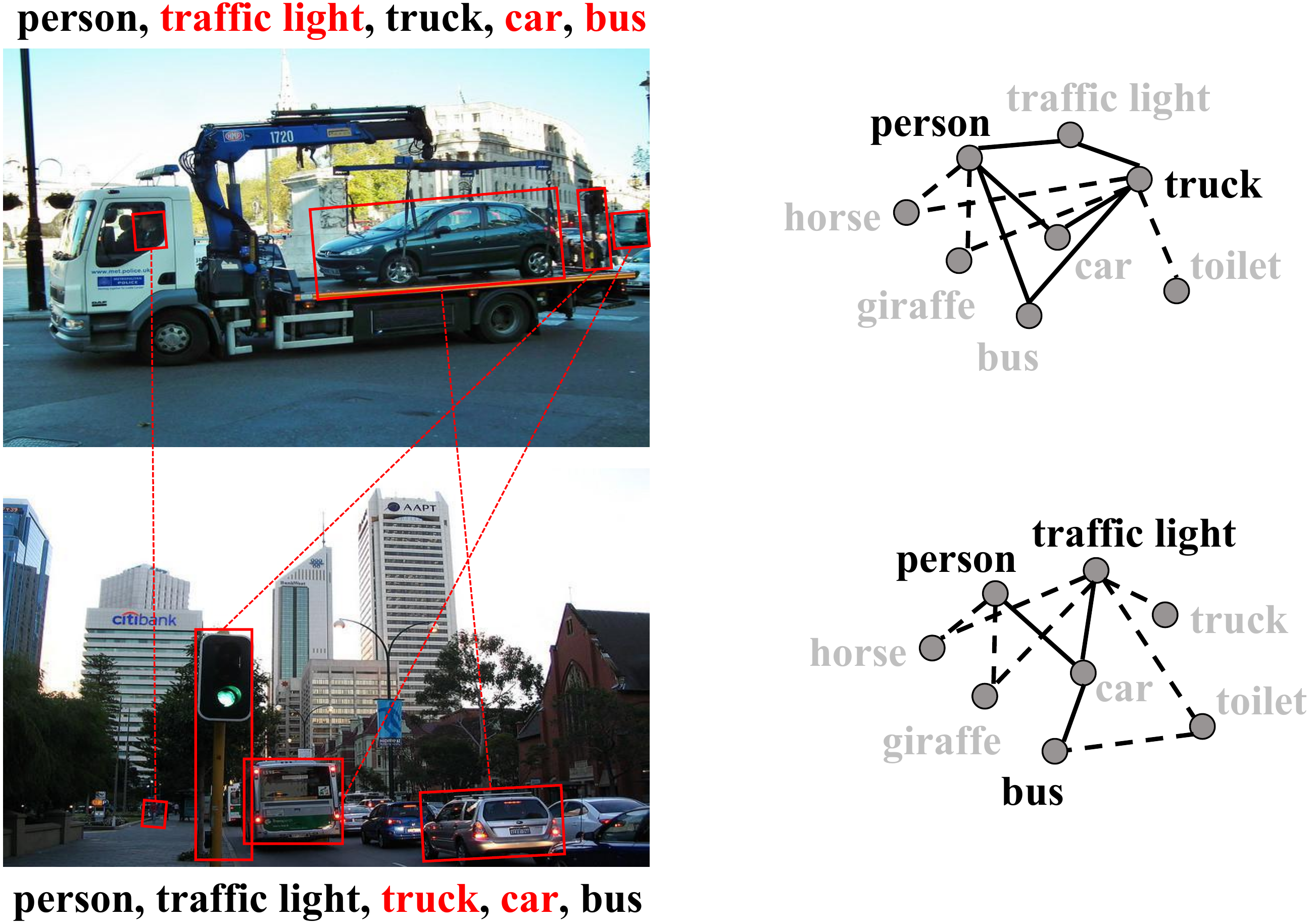}
   \vspace{-8pt}
   \caption{Two examples of images with partial labels (unknown labels are highlighted in red). We can mine the intra-image and cross-image correlations to help complement the unknown labels. }
   \vspace{-10pt}
   \label{fig:motivation}
   \vspace{-12pt}
\end{figure}

Current algorithms mainly consider multi-label recognition as a multiple binary classification task. Treating the unknown labels as missing or negative labels is an intuitive way to adapt these algorithms to address the MLR-PL task \cite{sun2017revisiting,joulin2016learning}. However, it results in an obvious performance drop as it loses some data or even incurs some noisy labels. Fortunately, strong semantic correlations within each image and cross different images exist, and these correlations can efficiently help to transfer semantic knowledge of known labels to construct the unknown labels: i) Label co-occurrences are widespread in real-world images, e.g., tables tend to co-occur with chairs and cars are likely to co-exist with roads; ii) Objects of the same category in different images may share similar visual appearances, and thus images with similar visual features may have the same labels. 

In this work, we explore mining these correlations to help complement the unknown labels by a novel structured semantic transfer (SST) framework. It consists of two complementary modules that learn image-specific co-occurrence to help transfer semantic labels within each image and category-specific feature similarities to transfer semantic labels across different images. Although previous work \cite{huynh2020interactive} also takes notice of label/image dependencies, it merely introduces statistical co-occurrence and image-level similarities to regularize training. Instead, the SST framework aims to learn fine-grained image-specific co-occurrence and category-specific feature similarities, which can help construct accurate pseudo labels for the unknown labels to facilitate the MLR-PL task. For example in Figure \ref{fig:motivation}, the feature vectors of \emph{truck} are similar in two different images and we can use the annotated \emph{truck} of the upper image to help complement the unknown \emph{truck} of the lower image. Similarly, \emph{traffic light} has high co-occurrence probability with \emph{car}, and we can complete this unknown label based on the co-occurrence.

The SST framework builds on a semantic-aware representation learning (SARL) module that incorporates category semantic to help learn category-specific feature representation. Then, an intra-image semantic transfer (IST) module is designed to learn a co-occurrence matrix among all categories for each image and map the known labels to complement some unknown labels based on the learned co-occurrences. Meanwhile, a cross-image semantic transfer (CST) module is introduced to measure the similarities of feature representations that belong to the same category and are from different images. It then transfers the semantic known labels to help complement some unknown labels with high similarity. Finally, the known labels and complemented labels are used to supervise training the multi-label recognition model. 

The contributions of this work are summarized into three folds. First, we introduce a structured semantic transfer framework to simultaneously mine intra-image and cross-image correlations to help complement the unknown labels. Second, two complementary modules (i.e., intra-image and cross-image semantic transfer) are incorporated to transfer semantic within each image and cross different images to generate pseudo labels accurately. Finally, we conduct extensive experiments on variant datasets to demonstrate the effectiveness of the proposed SST framework. We also perform ablative studies to analyze the  contribution of each module for better understanding. 
%We will release the codes and trained models for further research.

\begin{figure*}[!t]
   \centering
   \includegraphics[width=0.70\linewidth]{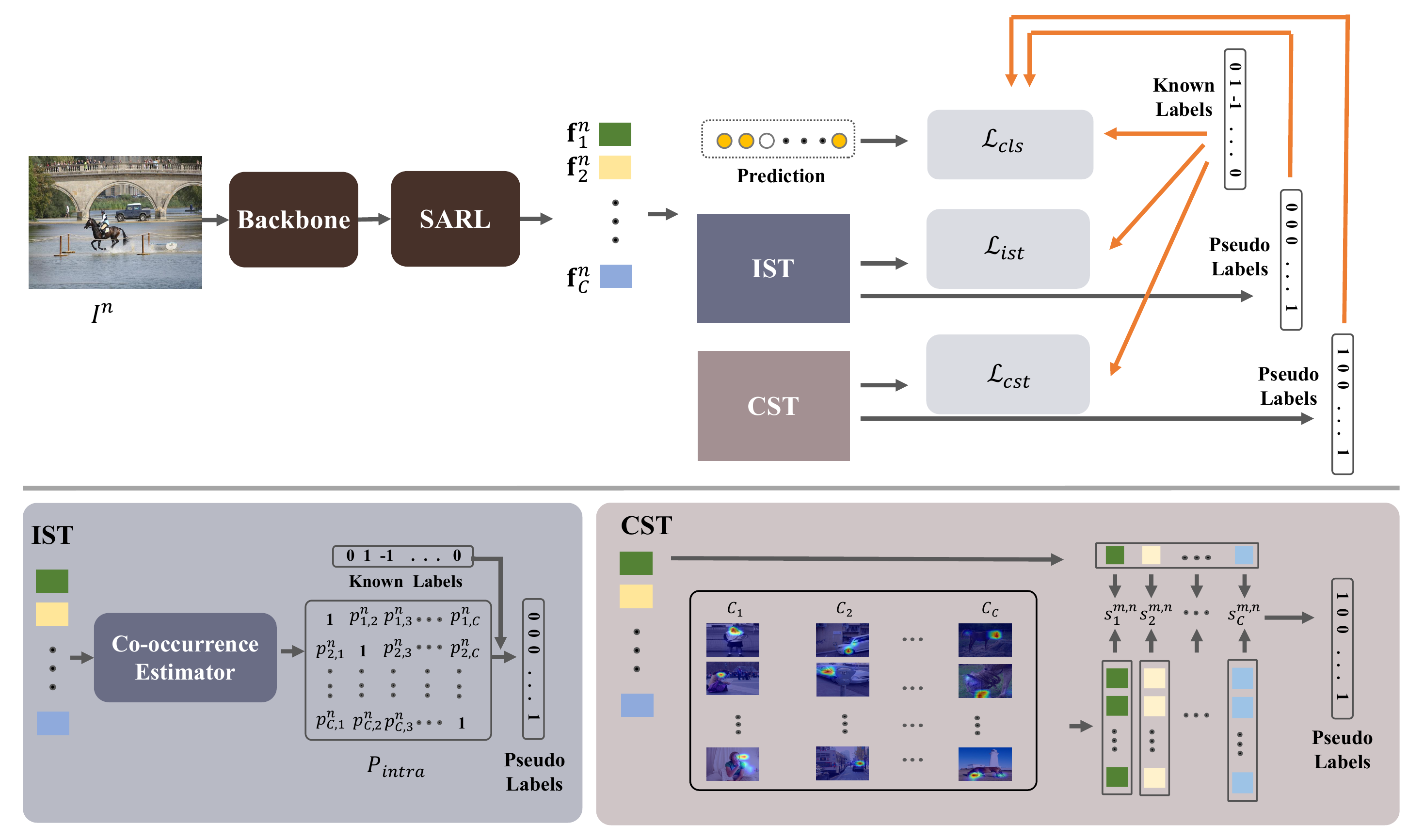}
   \vspace{-10pt}
   \caption{An overall illustration of the proposed structured semantic transfer framework. The upper part is the overall pipeline that consists of the IST and CST modules to generate pseudo labels, which are then fed to supervise training the multi-label recognition model. The lower part is the detailed implementations of the IST and CST modules. The IST module first predicts the label co-occurrence matrix and then maps the known labels to complement the unknown labels. The CST module first learns category-level feature similarities across different images and then also maps to generate the pseudo labels.}
   \vspace{-18pt}
   \label{fig:framework}
\end{figure*}

\vspace{-4pt}
\section{Related Works}
Multi-label image recognition receives increasing attention \cite{wei2016hcp,chen2020knowledge} since it is more practical and necessary than its single-label counterpart. To solve this task, lots of efforts are dedicated to discovering discriminative local regions for feature enhancement by object proposal algorithms \cite{wei2016hcp,yang2016exploit} or visual attention mechanisms \cite{ba2014multiple,chen2018recurrent}. Another line of works propose to capture label dependencies to regularize training multi-label recognition models and thus improve their performance \cite{wang2016cnn,wang2017multi,chen2019multi,chen2019learning}. These works either introduce the RNN/LSTM to implicitly capture label dependencies \cite{wang2016cnn,wang2017multi} or explicitly model the label dependencies in the form of structured graphs and exploit the graph neural networks \cite{li2016gated} to adaptively capture the label dependencies. Recently, Chen et al. \cite{chen2019learning} present state-of-the-art results on several multi-label datasets by using semantic decoupling to obtain semantic-aware features for different category labels, and we employ their semantic decoupling module for learning category-specific features in this work. However, despite achieving remarkable progress, all these methods rely on data-hungry deep neural networks \cite{simonyan2015very,he2016deep} to learn discriminative feature representation, and thus require large-scale and clean datasets (e.g., Visual Genome \cite{krishna2016vg}, MS-COCO \cite{lin2014microsoft} and Pascal VOC \cite{everingham2010pascal}) to train the deep neural networks. However, it is time-consuming and labor-intensive to annotate a complete list of labels for every image, making collecting large-scale and complete multi-label datasets less practical and scalable.

To reduce the annotation cost, some works propose to learn multi-label recognition models with partial labels, i.e., merely some labels are known \cite{durand2019learning,huynh2020interactive}. To deal with this task, some works \cite{bucak2011multi,wang2014binary,sun2017revisiting} simply regard the unknown labels as negative labels, and train the models with a similar scheme for the fully labeled setting. These methods could suffer from severe performance drop because many positive labels may be wrongly annotated as negative. Some other works \cite{tsoumakas2007multi} treat multi-label recognition as multiple independent binary classifications. However, it ignores the label dependencies that play a key role in multi-label recognition. To overcome this issue, some works exploit label dependencies to transfer the known labels to help complement the unknown label \cite{xu2013speedup,yu2014large}. Cabral et al. \cite{cabral2011matrix} introduce the low-rank regularization to exploit label correlations and complete unprovided labels, while Wu et al. \cite{WuLG15iccv} similarly adopts a low rank empirical risk minimization. A mixed graph is also utilized in \cite{WuLG15iccv} to encode a network of label dependencies. In \cite{KapoorVJ12nips}, missing labels are treated as latent variables in probabilistic models and predicted by posterior inference using Bayesian networks.Most of these works depend on solving an optimization problem that requires loading the whole training set, which cannot be integrated into deep networks for batch-level training. Such limitations result in inferior performance since fine-tuning is critical in transferring pre-trained DNN models. More recently, Durand et al. \cite{durand2019learning} propose a normalized BCE loss to exploit label proportion information and use it to train the model with partial labels. Huynh et al. \cite{huynh2020interactive} introduce statistical label co-occurrence and image-level feature similarity to regularize training networks.

Different from these methods, the proposed framework introduces two complementary modules, in which the first module learns image-specific label co-occurrence correlations to transfer provided labels within the same image to complement unknown labels and the second module learns category-level feature similarity correlations to transfer provided labels across different images to complement unknown labels. The two modules can be seamlessly incorporated into existing deep neural network models for multi-label recognition and trained in an end-to-end manner.

\vspace{-6pt}
\section{Structured Semantic Transfer}
In this section, we introduce the proposed SST framework that mines intra-image and cross-image correlations to help complement the unknown labels. It adopts a semantic-aware representation learning module to extract category-specific feature vectors for each image. The IST module first learns the co-occurrence probability of each category pair and then constructs a co-occurrence matrix for each image. It then transfers the semantic knowledge of the known labels to complement some unknown labels based on the learned co-occurrence matrix. Meanwhile, the CST module learns the similarities among feature vectors of the same category from different images. Similarly, we can also exploit the known labels to complement some unknown labels based on the learned similarities. In this way, we can obtain pseudo labels for unknown labels accurately and use both known labels and pseudo labels to train the multi-label models. An overall illustration is presented in Figure \ref{fig:framework}.

\noindent\textbf{Notation. }Here, we give an introduction to the notations used in the paper. We denote the training set as $\mathcal{D}=\{(I^1,\textbf{y}^1), ..., (I^N,\textbf{y}^N)\}$, in which $N$ is the number of training samples. $\textbf{y}^n=\{y^n_1, \cdots, y^n_C\}\in \{-1, 0, 1\}^C$ is the label vector of the $n$-th sample and $C$ is the label number. $y^n_c$ is assigned to 1 if label $c$ exists in the $n$-th image, assigned to -1 if it does not exist, and assigned to 0 if it is unknown. 

\vspace{-6pt}
\subsection{Semantic-Aware Representation Learning}
Given an input image $I^n$, we first utilize a backbone network to extract global feature maps $\textbf{f}^n$, and then follow recent work \cite{chen2019learning} to adopt the semantic decoupling module to learn semantic-aware representation of each category, denoted as $[\textbf{f}^n_1, \textbf{f}^n_2, \cdots, \textbf{f}^n_C]$. We use a gated graph neural network \cite{chen2019knowledge,chen2018knowledge,chen2021cross} and linear classifier followed by a sigmoid function to compute the probability score vectors $\textbf{p}^n$.

\vspace{-4pt}
\subsection{Intra-image Semantic Transfer}
There exist strong co-occurrence correlations among semantic labels in real-world images, and these correlations can effectively guide transferring semantic knowledge of known labels to generate pseudo labels for unknown labels. Current work \cite{huynh2020interactive} applies dataset-level statistical correlations to achieve this end. However, the statistical correlations are not appropriate for every image, and thus inevitably incur some incorrect labels. To avoid this problem, the IST module is proposed to learn the image-specific co-occurrence matrix and apply this matrix to complement unknown labels for the corresponding image. 

Given the semantic feature vectors $[\textbf{f}^n_1, \textbf{f}^n_2, \cdots, \textbf{f}^n_C]$ of input image $I^n$, we need to compute the co-occurrence probability for each category pair. For categories $i$ and $j$, we first concatenate the feature vector $\textbf{f}^n_i$ and $\textbf{f}^n_j$, and then feed the concatenated features to compute their co-occurrence probability, formulated as
\begin{equation}
 p_{i,j}^{n}=\phi_{intra}([\textbf{f}^n_i, \textbf{f}^n_j]),
\end{equation}
where $\phi_{intra}(\cdot)$ is implemented by several stacked fully connected layers. We compute the probabilities for all pairs and obtain a co-occurrence matrix $\textbf{P}^n_{intra}\in \mathcal{R}^{C\times C}$. Then, we estimate pseudo labels for unknown labels based on the co-occurrence matrix and known labels. For category $i$ that is not provided, we can compute its pseudo label by 
\begin{equation}
 \hat{y}^{n}_{i}=\textbf{1}[(\sum_{\{j|y^{n}_{j}=1\}}p_{i,j}^{n} \cdot y^{n}_{j}) \ge \theta_{intra}],
\end{equation}
where $\textbf{1}[\cdot]$ is an indicator function whose value is 1 if the argument is positive and is 0 otherwise. $\theta_{intra}$ is a threshold that helps to exclude the unlikely labels. We compute the pseudo labels for all unknown labels and combine it with known labels, obtaining $\hat{\textbf{y}}^{n}=\{\hat{y}^{n}_1,\hat{y}^{n}_2, \cdots, \hat{y}^{n}_C\}$.

Formally, the co-occurrence prediction can be considered as a binary classification task, and we can train it using the binary cross entropy (BCE) loss. However, it is very difficult to train the co-occurrence predictor because positive and negative pairs are extremely imbalanced. To address this task, we introduce the asymmetric loss \cite{ben2020asymmetric} that dynamically down-weights the importance of easy negative pair, defined as 
\begin{equation}
\mathcal{L}_{ist}=\sum_{n=1}^N \sum_{\{i,j\}}\ell_{i,j}^n,
\end{equation}
where
\begin{equation}
\ell_{i,j}^n=
    \begin{cases}
         (1-p_{i,j}^n)^{\gamma_1}\log(p_{i,j}^n) \quad &\{i,j\}\in \mathcal{D}^n\\
         (p_{i,j}^n-m)^{\gamma_2}\log(1-p_{i,j}^n)  \quad &\{i,j\}\notin \mathcal{D}^n.
    \end{cases}
\end{equation}
Here, $\mathcal{D}^n$ is the set of label pairs that co-occur in image $I^n$. $\gamma_1$, $\gamma_2$, and $m$ are the parameters to balance the loss and they are empirically set to 1, 2, and 0.05.

\subsection{Cross-image Semantic Transfer}
It is intuitive that the objects of the same category in different images share similar visual appearance. In other words, if two images share similar visual features, they tend to have the same labels. In the context of multi-label images, it is difficult to mine label correlation via image-level feature similarities. In this work, we design the CST module to learn category-level feature similarities and transfer known labels of images with high similarities to help complement unknown labels.

For each category $c$ of images $I^n$ and $I^m$, we use the cosine distance to compute their similarity, formulated as 
\begin{equation}
 s^{n,m}_c=cosine(\textbf{f}^n_c, \textbf{f}^m_c) = \frac{\textbf{f}^n_c\cdot\textbf{f}^m_c}{||\textbf{f}^n_c||\cdot||\textbf{f}^m_c||}.
\end{equation}
Suppose the label of category $c$ is missing in image $I^n$, we select image set $\mathcal{D}_c = \{m | y^{m}_{c}=1\}$, in which every image has positive label $c$. We first compute the average similarities $s^n_c$ between $\textbf{f}^n_c$ and the correspond feature vectors of the images in $\mathcal{D}_c$, and then estimate the existence of category $c$ by 
\begin{equation}
 \tilde{y}^{n}_{c}=\textbf{1}[(\frac{1}{|\mathcal{D}_c|}\sum_{\{m\in\mathcal{D}_c\}}s_{c}^{n,m} \cdot y^{m}_{c}) \ge \theta_{inter}].
\end{equation}
Similarly, $\textbf{1}[\cdot]$ is an indicator function and $\theta_{inter}$ is a threshold. We also estimate the pseudo labels for all unknown labels and combine it with known labels, obtaining $\tilde{\textbf{y}}^{n}=\{\tilde{y}^{n}_1, \tilde{y}^{n}_2, \cdots, \tilde{y}^{n}_C\}$.

It is expected that the similarity between $\textbf{f}^n_c$ and $\textbf{f}^m_c$ tends to be high if images $I^n$ and $I^m$ have the same positive label $c$, and the similarity should be low otherwise. Thus, it can be formulated as a ranking task and we introduce a pair loss for training, formulated as
\begin{equation}
\mathcal{L}_{cst}=\sum_{n=1}^N\sum_{m=1}^N\sum_{c=1}^C \ell^{n,m}_c,
\end{equation}
where
\begin{equation}
\ell^{n,m}_c=
    \begin{cases}
         1-s^{n,m}_c \quad & y_c^n=1,y_c^m=1\\
         1+s^{n,m}_c \quad & otherwise.
    \end{cases}
\end{equation}

\subsection{Optimization}
We follow previous work to use the partial binary cross entropy loss as objective function. Specifically, given the predicted probability distribution $\textbf{p}^n=\{p^n_1, p^n_2, \cdots p^n_C\}$ and the ground truth, the objective function can be defined as
\begin{equation}
\begin{aligned}
\ell(\textbf{p}^n, \textbf{y}^n)=&\frac{1}{\sum_{c=1}^C|y^n_c|}\sum_{c=1}^C[\textbf{1}(y^n_c=1)\log(p^n_c) \\
&+\textbf{1}(y^n_c=-1)\log(1-p^n_c)].
\end{aligned}
\end{equation}
We define similar objective functions for the pseudo labels generated by the intra-image and cross-image semantic transfer modules, i.e., $\ell(\textbf{p}^n, \hat{\textbf{y}}^{n})$ and $\ell(\textbf{p}^n, \tilde{\textbf{y}}^{n})$. And the final classification loss is defined as summing the three losses over all samples, formulated as
\begin{equation}
\begin{aligned}
\mathcal{L}_{cls}=&\sum_{n=1}^N(\ell(\textbf{p}^n, \textbf{y}^n)+\ell(\textbf{p}^n, \hat{\textbf{y}}^n) +\ell(\textbf{p}^n, \tilde{\textbf{y}}^n)).
\end{aligned}
\end{equation}
The final loss can be defined as summing up the classification loss, the intra-image and cross-image losses
\begin{equation}
\mathcal{L}=\mathcal{L}_{cls}+\lambda_{1}\mathcal{L}_{ist}+\lambda_{2}\mathcal{L}_{cst}.
\label{eq:total-loss}
\end{equation}
Here, $\lambda_{1}$ and $\lambda_{2}$ are the balance parameters that ensure the three losses have comparable magnitude, so that we set $\lambda_{1}$ and $\lambda_{2}$ to 10.0 and 0.05 in the experiments.

\section{Experiments}

\begin{table*}
  \centering
  \small
  \begin{tabular}{c|c|ccccccccc|c}
  \hline
  \centering Datasets & Methods & 10\% & 20\% & 30\% & 40\% & 50\% & 60\% & 70\% & 80\% & 90\% & Ave. mAP \\
  \hline
  \hline
  \vspace{2pt} \centering \multirow{6}*{MS-COCO} & SSGRL & 62.5 & 70.5 & 73.2 & 74.5 & 76.3 & 76.5 & 77.1 & 77.9 & 78.4 & 74.1 \\
  \centering ~ & GCN-ML & 63.8 & 70.9 & 72.8 & 74.0 & 76.7 & 77.1 & 77.3 & 78.3 & 78.6 & 74.4 \\
  \centering ~ & KGGR & 66.6 & 71.4 & 73.8 & 76.7 & 77.5 & 77.9 & 78.4 & 78.7 & 79.1 & 75.6 \\
  \centering ~ & Curriculum labeling & 26.7 & 31.8 & 51.5 & 65.4 & 70.0 & 71.9 & 74.0 & 77.4 & 78.0 & 60.7 \\
  \centering ~ & Partial BCE & 61.6 & 70.5 & 74.1 & 76.3 & 77.2 & 77.7 & 78.2 & 78.4 & 78.5 & 74.7 \\
  \centering ~ & Ours & \textbf{68.1} & \textbf{73.5} & \textbf{75.9} & \textbf{77.3} & \textbf{78.1} & \textbf{78.9} & \textbf{79.2} & \textbf{79.6} & \textbf{79.9} & \textbf{76.7} \\
  \hline
  \hline
  \centering \multirow{6}*{VG-200} & SSGRL & 34.6 & 37.3 & 39.2 & 40.1 & 40.4 & 41.0 & 41.3 & 41.6 & 42.1 & 39.7 \\
  \centering ~ & GCN-ML & 32.0 & 37.8 & 38.8 & 39.1 & 39.6 & 40.0 & 41.9 & 42.3 & 42.5 & 39.3 \\
  \centering ~ & KGGR & 36.0 & \textbf{40.0} & \textbf{41.2} & 41.5 & 42.0 & 42.5 & \textbf{43.3} & \textbf{43.6} & \textbf{43.8} & 41.5 \\
  \centering ~ & Curriculum labeling & 12.1 & 19.1 & 25.1 & 26.7 & 30.0 & 31.7 & 35.3 & 36.8 & 38.5 & 28.4 \\
  \centering ~ & Partial BCE & 27.4 & 38.1 & 40.2 & 40.9 & 41.5 & 42.1 & 42.4 & 42.7 & 42.7 & 39.8 \\
  \centering ~ & Ours & \textbf{38.8} & 39.4 & 41.1 & \textbf{41.8} & \textbf{42.7} & \textbf{42.9} & 43.0 & 43.2 & 43.5 & \textbf{41.8} \\
  \hline
  \hline
  \centering \multirow{6}*{Pascal VOC 2007} & SSGRL & 77.7 & 87.6 & 89.9 & 90.7 & 91.4 & 91.8 & 92.0 & 92.2 & 92.2 & 89.5 \\
  \centering ~ & GCN-ML & 74.5 & 87.4 & 89.7 & 90.7 & 91.0 & 91.3 & 91.5 & 91.8 & 92.0 & 88.9 \\
  \centering ~ & KGGR & 81.3 & 88.1 & 89.9 & 90.4 & 91.2 & 91.3 & 91.5 & 91.6 & 91.8 & 89.7 \\
  \centering ~ & Curriculum labeling & 44.7 & 76.8 & 88.6 & 90.2 & 90.7 & 91.1 & 91.6 & 91.7 & 91.9 & 84.1 \\
  \centering ~ & Partial BCE & 80.7 & 88.4 & 89.9 & 90.7 & 91.2 & 91.8 & 92.3 & 92.4 & 92.5 & 90.0 \\
  \centering ~ & Ours & \textbf{81.5} & \textbf{89.0} & \textbf{90.3} & \textbf{91.0} & \textbf{91.6} & \textbf{92.0 } & \textbf{92.5} & \textbf{92.6} & \textbf{92.7} & \textbf{90.4} \\
  \hline
  \end{tabular}
  \vspace{-8pt}
  \caption{Performance of our SST framework and current state-of-the-art competitors for MLR-PL on the MS-COCO, VG-200 and Pascal VOC 2007 datasets. The best results are highlighted in bold.}
  \vspace{-18pt}
  \label{tab:mAP-results}
\end{table*}

\subsection{Experimental Settings}

\noindent{\textbf{Datasets.}}\quad We follow previous works \cite{durand2019learning} to conduct experiments on the MS-COCO \cite{lin2014microsoft}, Visual Genome \cite{krishna2016vg}, and Pascal VOC 2007 \cite{everingham2010pascal} datasets for evaluation. MS-COCO contains about 120k images that cover 80 daily-life categories. It is further divided into a training set of about 80k images and a validation set of about 40k images. Visual Genome contains 108,249 images and covers 80,138 categories. Since most categories have very few samples, we merely consider the 200 most frequent categories, resulting in a VG-200 subset. We randomly select 10,000 images as the test set and the rest 98,249 images as the training set. Pascal VOC 2007 is the most widely used dataset for multi-label evaluation. It contains about 10k images from 20 object categories, which is divided into a trainval set of about 5,011 images and a test set of 4,952 images.

Because the three datasets are fully annotated, we randomly drop some labels to create the training set with partial labels. In this work, the proportion of dropped labels varies from 10\% to 90\%, resulting in 90\% to 10\% known labels.

\noindent{\textbf{Evaluation Metric.}}\quad For a fair comparison, we follow current works \cite{durand2019learning,huynh2020interactive} to adopt the mean average precision (mAP) over all categories for evaluation under different proportions of known labels. The proportions are set to 10\%, 20\%, $\cdots$, 90\%. And we also computer average mAP over all proportions. The overall and per-class precision, recall, F1-measure are also widely used to evaluate multi-label image recognition \cite{chen2019learning} and we also adopt these metrics for more comprehensive evaluation. We present these results in the supplementary material due to the page limit.

\noindent{\textbf{Implementation Details.}}\quad To fairly compare with existing algorithms, we follow previous works \cite{durand2019learning,chen2019learning} to adopt the 101-layers ResNet \cite{he2016deep} as the backbone to extract features $\textbf{f}^n$. Then, we use exactly the same decoupling module to learn category-specific semantic representation and gated graph neural network to learn contextualized category-specific feature vectors as \cite{chen2019learning}. The co-occurrence estimation function $\phi_{intra}(\cdot)$ is implemented by three fully-connected layers, in which the first layer maps 1024-dimension vector to 512 followed by the ReLU function, the second layer maps 512-dimension vector to 1,024 also followed by ReLU, and the last layer maps to a score that indicates the co-occurrence probability. 
The proposed framework is trained using the loss $\mathcal{L}$ as shown in Equation \ref{eq:total-loss}. The parameters of the ResNet-101 are initialized by those pre-trained on the ImageNet \cite{deng2009imagenet} dataset and the parameters of all other layers are initialized randomly. The model is trained using the ADAM algorithm \cite{kingma2015adam} with a batch size of 32, momentums of 0.999 and 0.9, and a weight decay of $5 \times 10^{-4}$. The original learning rate is set to 0.00001, and it is divided by 10 for every 10 epochs. It is trained with 20 epochs in total. During training, the input image is resized to 512$\times$512, and we randomly choose a number from \{512, 448, 384, 320, 256\} as the width and height to crop patch. Finally, the cropped patch is further resized to 448$\times$448. Then we perform randomly horizontal flip and perform normalization. $\theta_{intra}$ and $\theta_{inter}$ are two crucial parameters that control the accuracy of the generated pseudo labels. In the training process, the parameters are set to 1 during the first 5 epochs to avoid incurring any pseudo labels. Then, they are set to 0.95 at epoch 6 and are decreased by 0.025 for every epoch until they reach the minimum $\theta_{intra}$ and $\theta_{inter}$, respectively. Both the minimum $\theta_{intra}$ and $\theta_{inter}$ are set to 0.75 based on the experimental results. During inference, the intra-image and cross-image semantic transfer modules are removed, and the image is resized to 448$\times$448 for evaluation. 

\begin{table*}
  \centering
  \small
  \begin{tabular}{c|ccccccccc|c}
  \hline
  \centering  Methods & 10\% & 20\% & 30\% & 40\% & 50\% & 60\% & 70\% & 80\% & 90\% & Ave. mAP \\
  \hline
  \hline
  \centering SSGRL & 62.5 & 70.5 & 73.2 & 74.5 & 76.3 & 76.5 & 77.1 & 77.9 & 78.4 & 74.1 \\
  \centering Ours IST w/ stat & 55.3 & 62.3 & 65.9 & 70.3 & 71.8 & 72.7 & 73.5 & 74.6 & 75.2 & 69.1 \\
  \centering Ours IST & 64.1 & 71.3 & 74.5 & 75.9 & 77.2 & 77.7 & 78.2 & 78.8 & 79.1 & 75.2 \\
  \centering Ours IST w/o $L_{ist}$ & 61.9 & 70.9 & 73.2 & 75.0 & 76.3 & 76.8 & 77.6 & 78.2 & 78.6 & 74.3 \\
  \centering Ours CST & 64.2 & 72.5 & 74.4 & 76.2 & 77.1 & 77.9 & 78.4 & 78.9 & 79.3 & 75.4 \\
  \centering Ours CST w/o $L_{cst}$ & 63.0 & 71.7 & 73.8 & 74.4 & 76.3 & 76.9 & 77.6 & 78.3 & 78.6 & 74.5 \\
  \centering Ours w/ SAM & 67.8 & 73.2 & 75.3 & 77.5 & 78.3 & 78.6 & 79.0 & 79.4 & 79.7 & 76.5 \\
  \centering Ours & \textbf{68.1} & \textbf{73.5} & \textbf{75.9} & \textbf{77.3} & \textbf{78.1} & \textbf{78.9} & \textbf{79.2} & \textbf{79.6} & \textbf{79.9} & \textbf{76.7} \\
  \hline
  \end{tabular}
  \vspace{-5pt}
  \caption{Comparison of mAP of the baseline SSGRL, our framework merely using IST with statistical co-occurrence (Ours IST w/ stat), our framework merely using IST (Ours IST), our framework merely using IST without loss $L_{ist}$ (Ours IST w/o $L_{ist}$), our framework merely using CST (Ours CST), our framework merely using CST without loss $L_{cst}$ (Ours CST w/o $L_{cst}$), our framework using SAM instead of SD (Ours w/ SAM) and our framework (Ours) on the MS-COCO dataset.}
  \vspace{-15pt}
  \label{tab:ablation-results}
\end{table*}

\subsection{Comparison with the State-of-the-art algorithms}
To evaluate the effectiveness of the proposed SST framework, we compare it with the following algorithms that can be classified into three folds: 1) \textbf{SSGRL} \cite{chen2019learning}, \textbf{GCN-ML} \cite{chen2019multi} and \textbf{KGGR} \cite{chen2020knowledge} introduce graph neural networks to model label dependencies and they achieve state-of-the-art performance on the traditional multi-label image recognition task. We adapt these three methods to address the multi-label recognition with partial labels by replacing the loss with partial BCE loss while keeping other component unchanged. 2) \textbf{Curriculum labeling} \cite{durand2019learning} alternately labels the unknown labels with high evidence to update the training set and retrains the model with updated training set. We also treat it as a strong baseline to address this task. 3) \textbf{Partial BCE} \cite{durand2019learning} is the most-recent algorithm that is proposed to address this task. It introduces a normalized BCE loss to better exploit partial labels to train the multi-label models. We also include this algorithm for comparison. For fair comparisons, we adopt the same ResNet-101 network as backbone and follow exactly the same train/val split settings. 

\subsubsection{Performance on MS-COCO}
We present the comparison results on the MS-COCO dataset as shown in Table \ref{tab:mAP-results}. We find the traditional multi-label recognition methods SSGRL and GCN-ML can achieve competitive performance when the proportion of known labels is high (e.g., 70\%-90\%), but suffer from obvious performance drop when the proportion decreases. Partial BCE can achieve competing performance even when the proportion decreases to 30\%. By introducing the intra-image and cross-image correlations to generate pseudo labels, our SST framework obtains the best performance for all the settings of different proportions of known labels. Specifically, it obtains the mAPs of 68.1\%, 73.5\%, 75.9\%, 77.3\%, 78.1\%, 78.9\%, 79.2\%, 79.6\%, 79.9\% on the settings of 10\%-90\% known labels, outperforming the second-best KGGR algorithm by 1.5\%, 2.1\%, 2.1\%, 0.6\%, 0.6\%, 1.0\%, 0.8\%, 0.9\%, 0.8\%, respectively. It is worth noting that SST can achieve more obvious performance improvement when the known labels are small, e.g., mAP improvement of 1.5\%, 2.1\% when the known label proportions are 10\% and 20\%. 

\subsubsection{Performance on VG-200}
VG-200 is a more challenging dataset that covers a lot more categories, and we also present the comparison results. As shown in Table \ref{tab:mAP-results}, our SST framework obtains the best performance over all proportion settings. Specifically, its average mAP is 41.8\%, outperforming the second-best KGGR algorithm by 0.3\%. Besides, it outperforms leading multi-label methods SSGRL and GCN-ML by 4.2\% and by 6.8\% when known labels are 10\%.

\subsubsection{Performance on Pascal VOC 2007}
Pascal VOC is the most-widely used dataset for multi-label recognition and we also present the results in Table \ref{tab:mAP-results}. As this dataset merely covers 20 categories and it is more simple than Visual Genome and MS-COCO, current algorithms achieve comparable results when keeping a certain proportion of known labels (e.g., more than 40\%). But their performances drop dramatically when the proportion decreases to 10\% and 20\%. Our SST framework also suffers performance drop, but it consistently outperforms current methods for all proportion settings. Specifically, it outperforms multi-label methods (i.e., SSGRL and GCN-ML) by 3.8\% and by 7.0\% and Partial BCE by 0.8\% when known labels are merely 10\%.

\vspace{-2pt}
\subsection{Ablative studies}
As discussed above, SSGRL can be treated as the baseline method and we stress the comparison with SSGRL to verify the contribution of the structured semantic transfer module. As shown in Table \ref{tab:ablation-results}, the SSGRL obtains an average mAP of 74.1\%. By introducing the structured semantic transfer module to complement the unknown labels, SST boosts the average mAP to 76.7\%, with an improvement of 2.6\%. And SST also performs consistently better than the baseline SSGRL method on different proportion settings.

The SST framework depends on the semantic-aware representation learning (SARL) module to learn semantic-aware representation. In this work, we use the semantic decoupling algorithm proposed in \cite{chen2019learning} to implement this module, because it achieves the state-of-the-art performance for the multi-label recognition task. It is noteworthy that we can also use other algorithms for learning semantic-aware representation. To verify this point, we replace the semantic decoupling algorithm with the semantic attention module (SAM) proposed in \cite{Ye2020ADD-GCN} to generate category-specific representation. As shown in Table \ref{tab:ablation-results}, ``Ours w/ SAM'' can also achieve comparable performance, suggesting the universality of the proposed SST framework.

% Besides the semantic decoupling module, \cite{Ye2020ADD-GCN} design a semantic attention module (SAM) to learn category-specific semantic representation, thus we replace the SD module with the SAM module. As presented in Table \ref{tab:ablation-results}, the proposed SST framework with SAM still receives comparative performance on all known labels proportion settings, that demonstrates our proposed framework is not limited to a specific extracting semantic method and the effectiveness of the SST framework does not depend on that of SSGRL.

% The foregoing comparisons with the baseline method demonstrate the effectiveness of the SST framework as a whole.

Since SST consists of two complementary modules, i.e., intra-image semantic transfer (IST) and cross-image semantic transfer (CST) modules, in the following we will conduct more ablative experiments to analyze the separate contributions of these two modules in detail.

\subsubsection{Analysis of intra-image semantic transfer (IST)}

\begin{figure}[!t] 
  \centering    
  \subfigure {
  \includegraphics[width=0.45\linewidth]{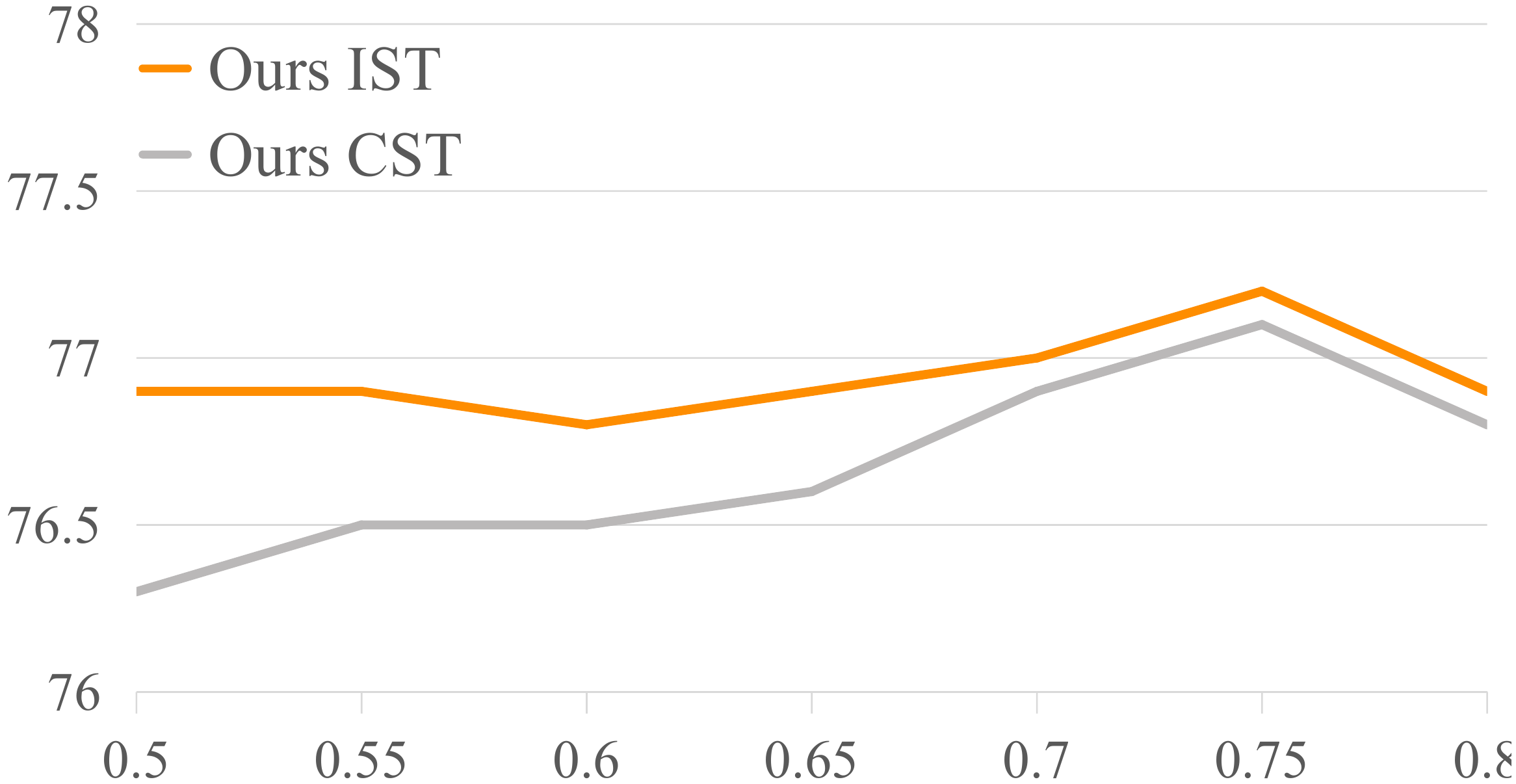}  
  }~     
  \subfigure {
  \includegraphics[width=0.45\linewidth]{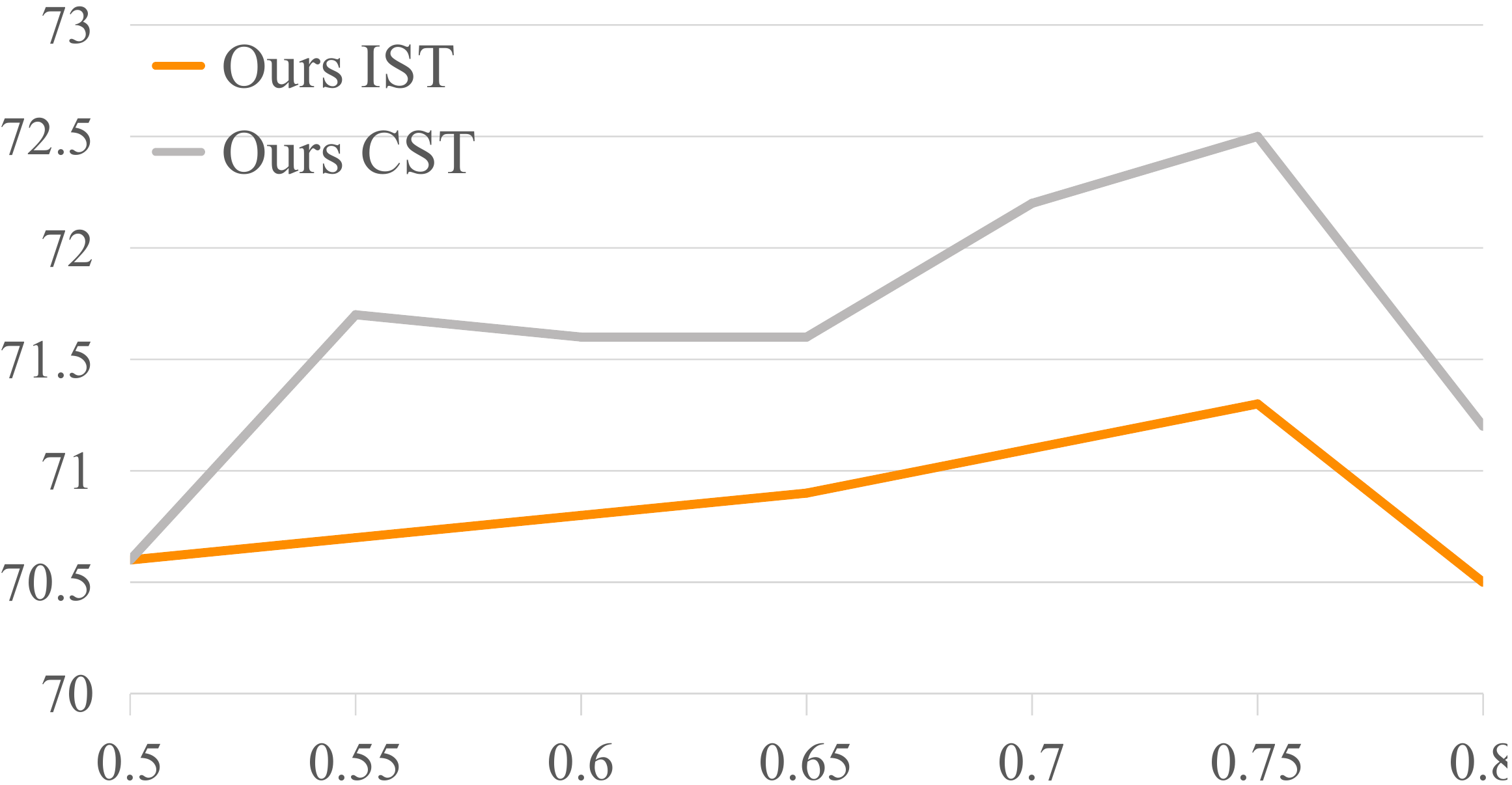}     
  }~    
  \vspace{-8pt}
  \caption{The performance of ``Ours IST'' and ``Ours CST'' with different  minimum thresholds on the 50\% (left) and 20\% (right) known labels settings.}     
  \vspace{-18pt}
  \label{fig:threshold}     
\end{figure}

\noindent\textbf{\\Effects of threshold $\theta_{intra}$.} $\theta_{intra}$ is a crucial threshold that controls the accuracy and recall of the pseudo labels. Setting it to a small value may recall some false positive labels while setting it to a large value may miss some true positive labels. We conduct experiments with minimum $\theta_{intra}$ varying from 0.5 to 0.8, and present the performance change on the settings with 20\% and 50\% known labels. As shown in Figure \ref{fig:threshold}, it shows that decreasing minimum $\theta_{intra}$ from 0.8 to 0.75 leads to performance improvement but further decreasing it leads to performance drop on both two settings. Thus, we set the minimum $\theta_{intra}$ as 0.75.

%\begin{figure}[!t]
%   \centering
%   \includegraphics[width=\linewidth]{./figures/threshold.pdf}
%   \vspace{-18pt}
%   \caption{The performance of ``Ours IST'' and ``Ours CST'' with different  minimum thresholds on the 50\% (left) and 20\% (right) known labels settings.}
%   \vspace{-18pt}
%   \label{fig:threshold}
%\end{figure}

\begin{figure}[!t]
   \centering
   \includegraphics[width=0.8\linewidth]{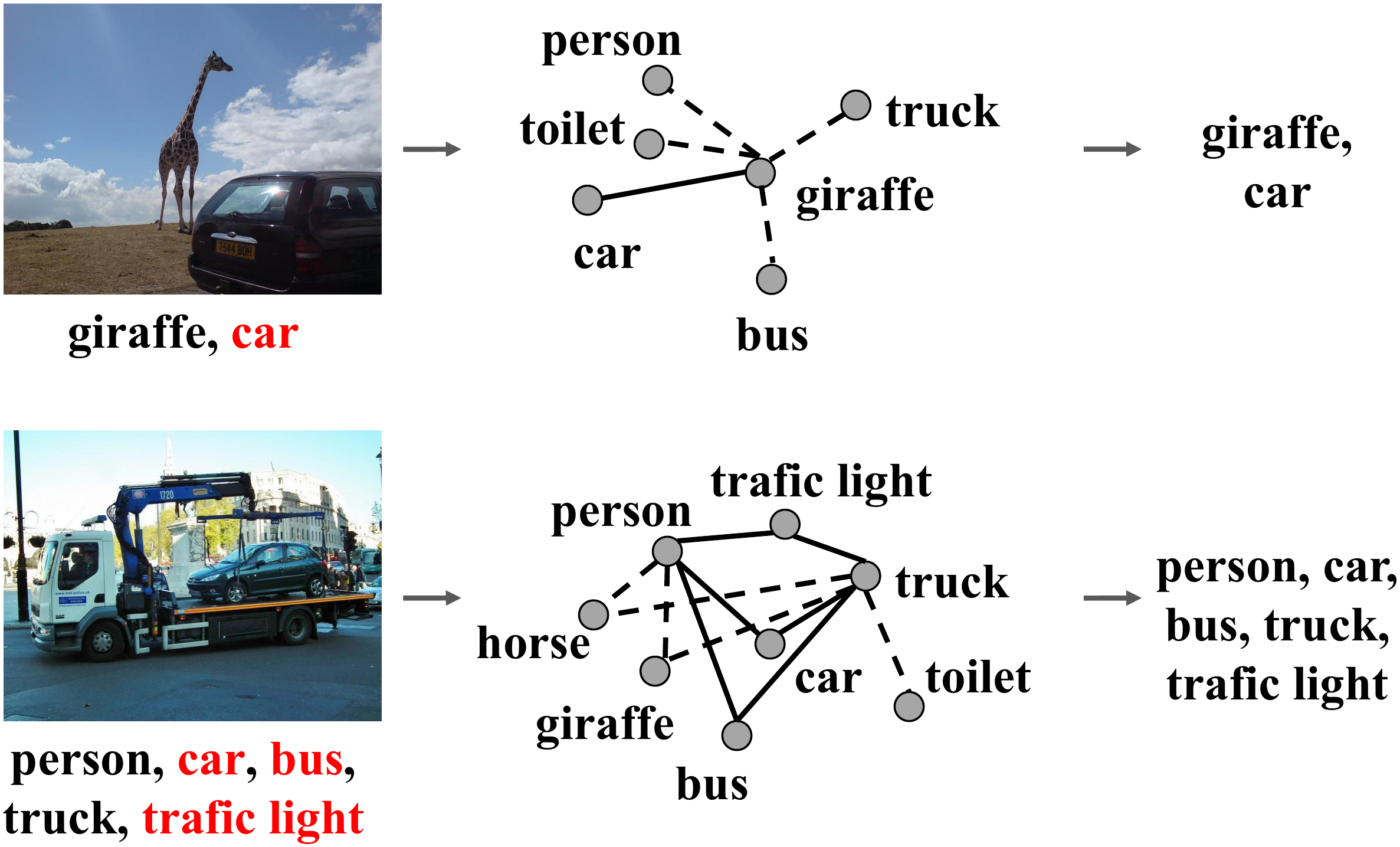}
   \vspace{-5pt}
   \caption{Examples of the image-specific co-occurrence matrices and the complemented labels: input images and labels (left), partial graphs of within-image co-occurrence (middle), and pseudo positive labels (right). The missing labels are highlighted in red. Two categories that have high co-occurrence probability are connected by solid line, otherwise, connected by a dotted line.}
   \label{fig:vis-ist}
   \vspace{-15pt}
\end{figure}

\noindent\textbf{Contribution of the IST module.} We then evaluate the actual contribution of the IST module by comparing the performance with and without this module. As shown in Table \ref{tab:ablation-results}, ``Ours IST'' means we merely use the IST module to generate pseudo labels. We find it achieves obvious mAP improvement compared with the baseline SSGRL, i.e., an average mAP improvement of 1.1\%. Besides, the loss $L_{ist}$ helps to learn accurate co-occurrence matrix. To evaluate its effectivenss, we conduct experiments that remove this loss for comparison (namely Ours IST w/o $L_{ist}$). As shown in Table \ref{tab:ablation-results}, it further decreases the average mAP by 0.9\%. 

Here, we learn image-specific co-occurrence matrix to generate pseudo labels. To demonstrate its effectiveness, we perform experiments that use statistical co-occurrence matrix computed on the training dataset to generate the pseudo labels, namely ``Ours IST w/ stat''. As shown in Table \ref{tab:ablation-results}, it suffers from dramatic performance drop. Specifically, the average mAP is merely 69.1\%, worse than that using image-specific co-occurrence matrix by 6.1\% in average mAP. One reason for this phenomenon is that statistical co-occurrence is not suitable for every image and thus it may incur many false positive labels for the unsuitable images.

\begin{figure}[!t]
   \centering 
   \includegraphics[width=0.8\linewidth]{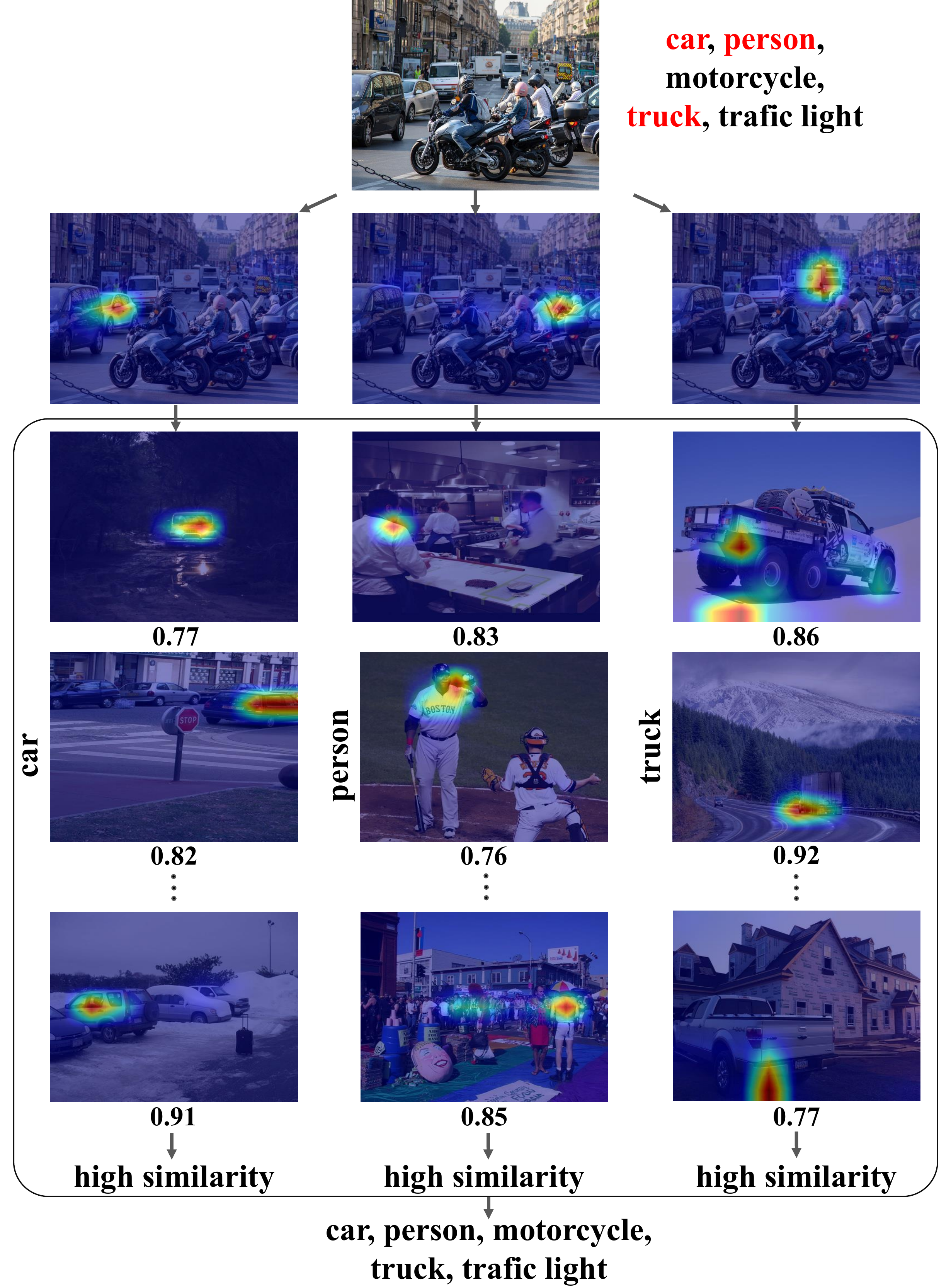}
   \vspace{-5pt}
   \caption{An example of category-specific feature similarities and the complemented labels: input image and labels (top), category-specific feature vectors (middle top), category-specific feature vectors of other images with known labels that are missing for the given image (middle in box), and the generated pseudo labels (bottom). The missing labels are highlighted in red.}
   \label{fig:vis-cst}
   \vspace{-15pt}
\end{figure}

To delve deep into the IST module, we visualize some examples of the image-specific co-occurrence matrices and how these matrices generate the pseudo labels in Figure \ref{fig:vis-ist}. As shown, it can capture the category pairs that frequently co-occur, like \textit{car} and \textit{person} in the second example. It can also assign a high co-occurrence probability to the pairs that rarely co-exist, e.g., \textit{giraffe} and \textit{car} in the first example. This also suggests that learning image-specific co-occurrences can better capture label correlations for each image and thus facilitate generating more accurate pseudo labels.

\subsubsection{Analysis of cross-image semantic transfer (CST)}
 
\noindent\textbf{\\Effects of threshold $\theta_{inter}$.} $\theta_{inter}$ performs similar role with $\theta_{inter}$ but it controls the generated labels by the CST module. Here, we also perform experiments that vary the minimum $\theta_{inter}$ from 0.5 to 0.8 on the settings with 20\% and 50\% known labels. The results are presented in Figure \ref{fig:threshold}. The mAP increases from 71.2\% to 72.5\% and from 76.8\% to 77.1\% on the settings of 20\% and 50\% known labels when decreasing minimum $\theta_{inter}$ from 0.8 to 0.75, and it drops obviously if further decreasing minimum $\theta_{inter}$. Thus, the minimum $\theta_{inter}$ is set to 0.75 in the experiments.

\noindent\textbf{Contribution of the CST module.} In this section, we add the CST module to the baseline SSGRL, namely ``Ours CST'', and compare it with baseline method to verify the contribution of CST. As shown in Table \ref{tab:ablation-results}, it shows that adding the CST module improves the average mAP from 74.1\% to 75.4\%, an improvement of 1.3\%. In this module, the loss $L_{cst}$ plays an important role in learning category-specific feature similarity. Here, we also evaluate its contribution by conducting an experiment that removes this loss (namely Ours CST w/o $L_{cst}$). It is observed that the average mAP decreases from 75.4\% to 74.5\%.

As discussed above, the CST module measures category-level feature similarity of the same category from different images to help complement the unknown labels. Here, we also visualize an example that loses labels of \emph{car}, \emph{person}, and \emph{truck} (Figure \ref{fig:vis-cst}). We can see that the features belonging to the same category but from different image share very high similarities, which help to recall the missing labels.

\section{Conclusion}
In this work, we aim to address the multi-label image recognition with partial labels task by designing a novel structured semantic transfer framework, which consists of an intra-image semantic transfer module that mines image-specific label co-occurrences and a cross-image semantic transfer module that mines category-level feature similarities, to transfer semantics of known labels to complement unknown labels for model training. We conduct extensive experiments on various multi-label datasets (e.g., MS-COCO, VG-200, and Pascal VOC) to demonstrate its superiority.

\section{Acknowledgements}
This work was supported by National Natural Science Foundation of China (No. 61876045, 61836012 and 62002069), the Natural Science Foundation of Guangdong Province (No. 2017A030312006) and Guangdong Provincial Basic Research Program (No. 102020369).

{\small
\bibliography{aaai22}
}

\clearpage

\section{Supplementary Experiments}

In the orginal paper, we have presented the mAP comparison with the state-of-the-art algorithms. Besides the mAP, the overall precision, recall, F1-measure (OP, OR, OF1) and per-class precision, recall, F1-measure (CP, CR, CF1) are also widely used for evaluating the multi-label image recognition task \cite{chen2018recurrent,wei2016hcp}. In this work, we have also adopted these metrics for more in-depth and comprehensive comparisons. Due to the page limit of the paper, we present these comparison results in this supplementary material.

\subsection{Evaluation Metrics}
Formally, the CP, CR, CF1, OP, OR, OF1 can be computed by
\begin{gather}
 OP=\frac{\sum_{i}{N^c_i}}{\sum_{i}{N^p_i}}, CP=\frac{1}{C} \sum_{i}{ \frac{N^c_i}{N^p_i} } \\
 OR=\frac{\sum_{i}{N^c_i}}{\sum_{i}{N^g_i}}, CR=\frac{1}{C} \sum_{i}{ \frac{N^c_i}{N^g_i} } \\
 OF1=\frac{2 \times OP \times OR }{OP+OR},  CF1=\frac{2 \times CP \times CR }{CP+CR}
\end{gather}
where $N^c_i$ is the number of images that are correctly predicted for the $i$-th label, $N^p_i$ is the number of predicted images for the $i$-th label, $N^g_i$ is the number of ground truth images for the $i$-th label. We also average the OP, OR, OF1, CP, CR, CF1 on all known label proportion settings.

\begin{table*}[b]
  \centering
  \begin{tabular}{c|c|cccccc}
  \hline
  \centering Datasets & Methods& Avg. OP & Avg. OR & Avg. OF1 & Avg. CP & Avg. CR & Avg. CF1 \\
  \hline
  \hline
  \centering \multirow{6}*{MS-COCO} & SSGRL & 86.3 & 64.8 & 73.9 & 82.1 & 58.4 & 68.1 \\
  \centering ~ & GCN-ML & 85.2 & 64.2 & 73.1 & 81.8 & 58.9 & 68.4 \\
  \centering ~ & KGGR & 84.0 & 65.6 & 73.7 & 81.4 & 60.9 & 69.7 \\
  \centering ~ & Curriculum labeling & \textbf{87.8} & 51.0 & 61.9 & 60.9 & 40.4 & 48.3 \\
  \centering ~ & Partial BCE & 86.7 & 64.7 & 74.0 & \textbf{83.1} & 58.9 & 68.8 \\
  \centering ~ & Ours & 86.3 & \textbf{67.7} & \textbf{75.8} & 82.8 & \textbf{62.6} & \textbf{71.2} \\
  \hline
  \hline
  \centering \multirow{6}*{VG-200} & SSGRL & 69.9 & 25.9 & 37.8 & 45.3 & 18.3 & 26.1 \\
  \centering ~ & GCN-ML & 64.1 & 28.2 & 38.7 & 44.6 & 18.2 & 25.6 \\
  \centering ~ & KGGR & 64.5 & \textbf{30.5} & \textbf{41.2} & \textbf{54.8} & \textbf{25.8} & \textbf{33.6} \\
  \centering ~ & Curriculum labeling & 66.4 & 15.4 & 23.6 & 20.4 & 7.6 & 10.9 \\
  \centering ~ & Partial BCE & 69.7 & 24.6 & 36.1 & 44.3 & 18.1 & 25.7 \\
  \centering ~ & Ours & \textbf{69.9} & 27.9 & 39.9 & 49.8 & 22.3 & 30.8 \\
  \hline
  \hline
  \centering \multirow{6}*{Pascal VOC 2007} & SSGRL & 91.2 & 84.4 & 87.7 & 87.8 & 81.4 & 84.5 \\
  \centering ~ & GCN-ML & 92.2 & 83.0 & 87.3 & \textbf{89.7} & 80.1 & 84.6 \\
  \centering ~ & KGGR & 90.5 & 82.9 & 86.5 & 88.5 & 81.4 & 84.7 \\
  \centering ~ & Curriculum labeling & \textbf{92.7} & 78.2 & 83.8 & 79.5 & 71.7 & 75.4 \\
  \centering ~ & Partial BCE & 91.8 & 84.3 & 87.9 & 88.8 & 81.3 & 84.8 \\
  \centering ~ & Ours & 91.3 & \textbf{85.3} & \textbf{88.2} & 88.3 & \textbf{83.0} & \textbf{85.6} \\
  \hline
  \end{tabular}
  \vspace{-2pt}
  \caption{The average OP, OR, OF1 and CP, CR, CF1 of the proposed SST framework and current state-of-the-art competitors for multi-label recognition with partial labels on the MS-COCO, VG-200 and Pascal VOC 2007 datasets. The best results are highlighted in bold.}
  \label{tab:results-detail}
  \vspace{-15pt}
\end{table*}

\subsection{Performance Analysis}
Similar to the original paper, we compare the proposed SST framework with the  SSGRL \cite{chen2019learning}, GCN-ML \cite{chen2019multi}, KGGR \cite{chen2020knowledge}, Curriculum labeling \cite{durand2019learning}, Partial BCE \cite{durand2019learning} on the metrics of average OP, OR, OF1, CP, CR, CF1 of the MS-COCO \cite{lin2014microsoft}, VG-200 \cite{krishna2016vg}, and Pascal VOC 2007 \cite{everingham2010pascal} datasets. The comparison results are presented in Table \ref{tab:results-detail}. As shown, the proposed SST framework achieves the best overall performance on these metrics over all current algorithms on all the three datasets. On MS-COCO, it achieves the OF1 and CF1 of 75.8\% and 71.2\%, outperforming the previous state-of-the-art KGGR algorithm by 2.1\% and 1.5\%.  Similarly, as Pascal VOC 2007 is a much more simple dataset and current algorithm can also achieve quite competing performance. As shown, the current best-performing Partial BCE algorithm achieves the average OF1 and CF1 of 87.9\% and 84.8\%. Still, the SST performs better than this algorithm, improving the OF1 and CF1 to 88.2\% and 85.6\%, respectively. These comparisons again demonstrate the superiority of the proposed SST framework.

\end{document}